# Evaluating Urbanization from Satellite and Aerial Images by means of a statistical approach to the texture analysis

**Amelia Carolina Sparavigna**

Department of Applied Science and Technology, Politecnico di Torino, Torino, Italy

**Abstract:** Statistical methods are usually applied in the processing of digital images for the analysis of the textures displayed by them. Aiming to evaluate the urbanization of a given location from satellite or aerial images, here we consider a simple processing to distinguish in them the "urban" from the "rural" texture. The method is based on the mean values and the standard deviations of the colour tones of image pixels. The processing of the input images allows to obtain some maps from which a quantitative evaluation of the textures can be obtained.

## 1. Introduction

The processing of images by means of a statistical approach can be rather simple, in particular when it is mainly based on quantities such as the mean values and the standard deviations [1]. Besides for its simplicity, this approach is valuable because it allows a very fast elaboration of images, being therefore quite useful when we need to analyse a large number of images. We used it, for instance, for investigating the texture transitions in liquid crystals [2-4]. Here we modify the method to have it suitable for the study of satellite and aerial images, from which we can measure the evolution of the local urbanization. Some examples of the method, on sites in Italy and in England, will be given using the images we can find in Google Earth.

In fact, the Google Earth software features historical data on several places around the world. We have, among the software icons, the clock icon that we can select to view the historical imagery in a time series. It is enough to select the icon and use the given sliding tool to view the progress of the selected location through time. Using Google Earth and its gallery of images, we can time travel to see the natural evolution of a place or the development of urban areas [5-7]. Let us note that the software is archiving, besides the satellite images, also some aerial imagery. In particular we have a large part of England and some of the Italian towns, covered by aerial images of 1943 or 1945. In this manner we can easily evaluate the processes linked to urbanization of such covered areas, after the World War II.

In the following section we will illustrate the method and give some examples. We will see that it is possible to distinguish the "urban" texture from the "rural" texture in the images. In fact, from the input images we are obtaining the maps which allow a quantitative evaluation for distinguishing the textures in them.

## 2. The statistical method

The image analysis method we use here can work on any colour image. However, instead of considering a full RGB picture, let us involve for computation the corresponding grayscale image. The starting point is therefore a RGB image of $N_x \times N_y$ pixels, reppresented by the three-channel brightness function:

$$b : I \to B, \quad I = [1, N_x] \times [1, N_y] \subset \mathsf{N}^2, \quad B = [0,255]^3 \subset \mathsf{N}^3.$$

After, we consider a grey-tone map obtained by giving $\widetilde{\beta}(i,j) = \frac{1}{3}\sum_{c=1}^{3} b_c(i,j)$, where $c$ corresponds to the three channels R, G, B. The integer indeces $i$ and $j$ are ranging in the $x$ and $y$ direction of the cartesian frame, corresponding to the image frame. Let us fix an integer $P$; we subdivide the image in squares having size $P \times P$, or, more specifically $P_x \times P_y$ with $P_x = P_y = P$. The upper left corrner of each

square is a pixel of the image that we indicate as $(i_P, j_P)$. We compute the local average brightness as:

$$M(i_P, j_P) = \frac{1}{P_x P_y} \sum_{\mu=1}^{P_x} \sum_{\nu=1}^{P_y} \widetilde{b}(i_P + \mu, j_P + \nu),$$

Let us define also the standard deviation:

$$\Sigma(i_P, j_P) = \sqrt{\frac{1}{P_x P_y} \sum_{\mu=1}^{P_x} \sum_{\nu=1}^{P_y} \left[ \widetilde{b}(i_P + \mu, j_P + \nu) - M(i_P, j_P) \right]^2}$$

Then, we evaluate the ratio:

$$\xi(i_P, j_P) = \frac{\Sigma(i_P, j_P)}{M(i_P, j_P)}$$

Among the values $\xi(i_P, j_P)$ we determine the largest one, $\Xi$. Then, we define a colour tone accordingly:

$$\xi_c(i_P, j_P) = \frac{255 \times \xi(i_P, j_P)}{\Xi} \qquad (*)$$

Then, from the original image we create a map made of squares with size $P \times P$ having a specific colour tone given by (*). Let us stress that, in each square, the pixels have the same colour tone (*).
We have a bright square, when the local standard deviation is large. This happens when the pixels of the original image in the given square have oscillating values. In the case of satellite images, this corresponds to an area containing houses and streets. The dark squares are coming from the large areas in the original images, which have more homogenous grey tones. These areas are corresponding to fields for agriculture. Let us see some examples.

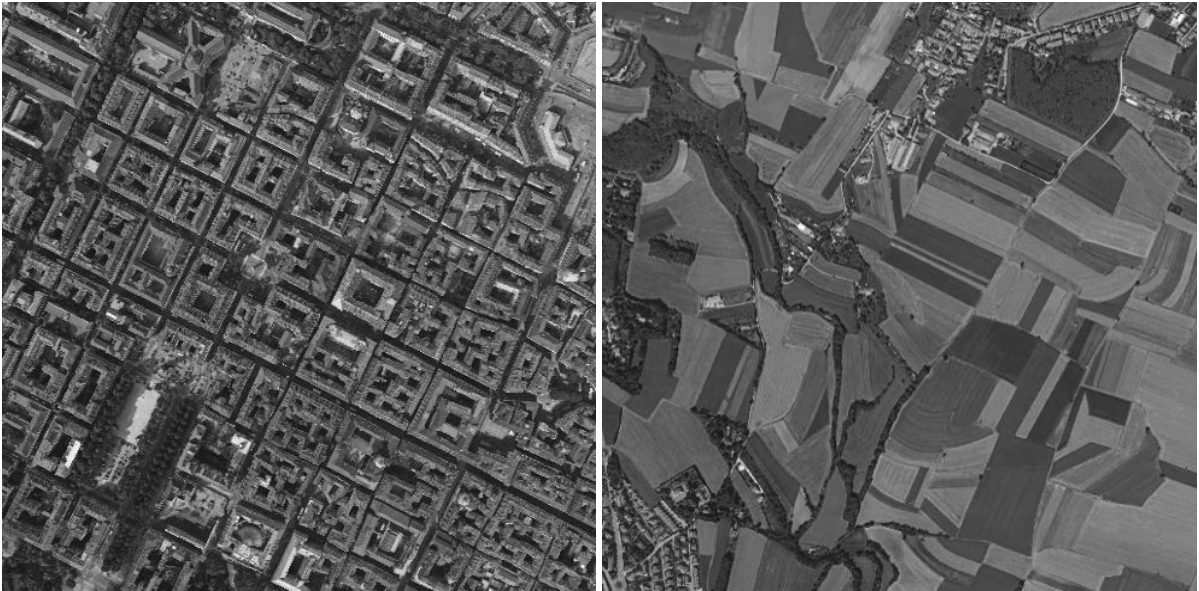

Figure 1: Urban area and agricultural land (Courtesy Google Earth).

## 3. Torino and agricultural land near the town

Let us consider two images from Google Earth (Figure 1). On the left we can see the central part of Torino, and on the right, some fields near the town. Let us apply the statistical method previously discussed. We have the following two maps given in the Figure 2.

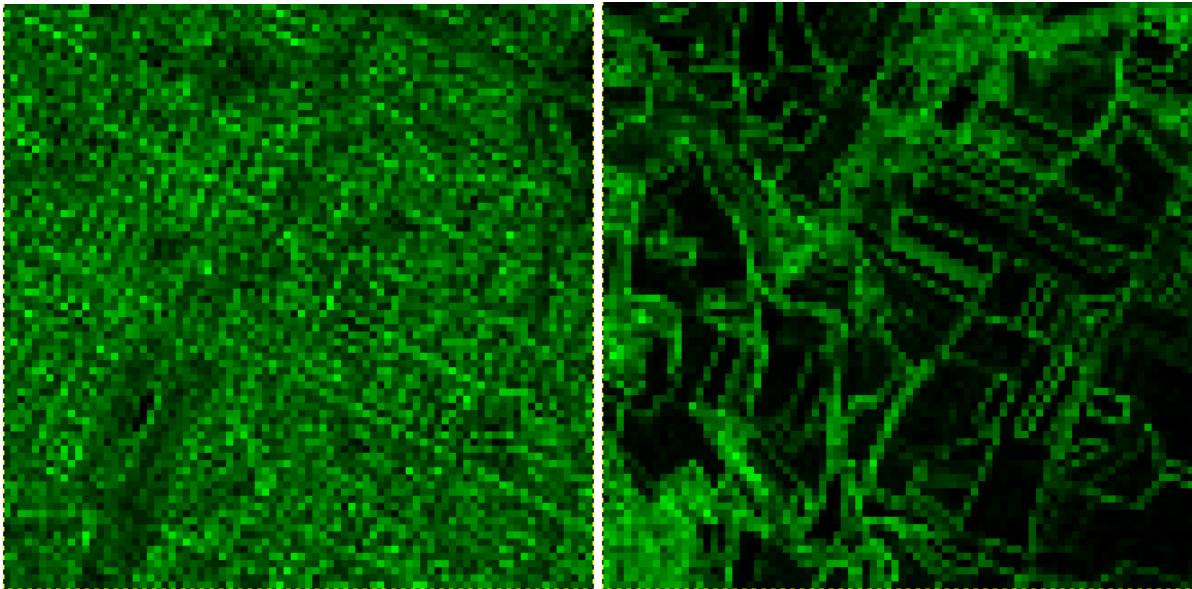

Figure 2: The maps made of squares obtained from the panels of Figure 1, according to the statistical method previously discussed.

As we can see from the map on the right in Fig.2, there are also evidenced some of the edges among fields. The homogenous areas turn into black areas. The two maps are evidencing the different textures of urban and rural areas: the urban areas have a large number of bright squares, homogeneously distributed, whereas the rural land has a large number of dark squares.

Counting the percentage of the bright squares can be a possible method for a quantitative measure of urbanization. Let us stress that the proposed approach is different from the simple pixelation we can obtain from a software such as GIMP, the GNU Image Manipulation Program. The GIMP pixelation of the panels in the Figure 1 are given in Figure 3. The GIMP pixelation is not useful for a quantitative evaluation of urbanization.

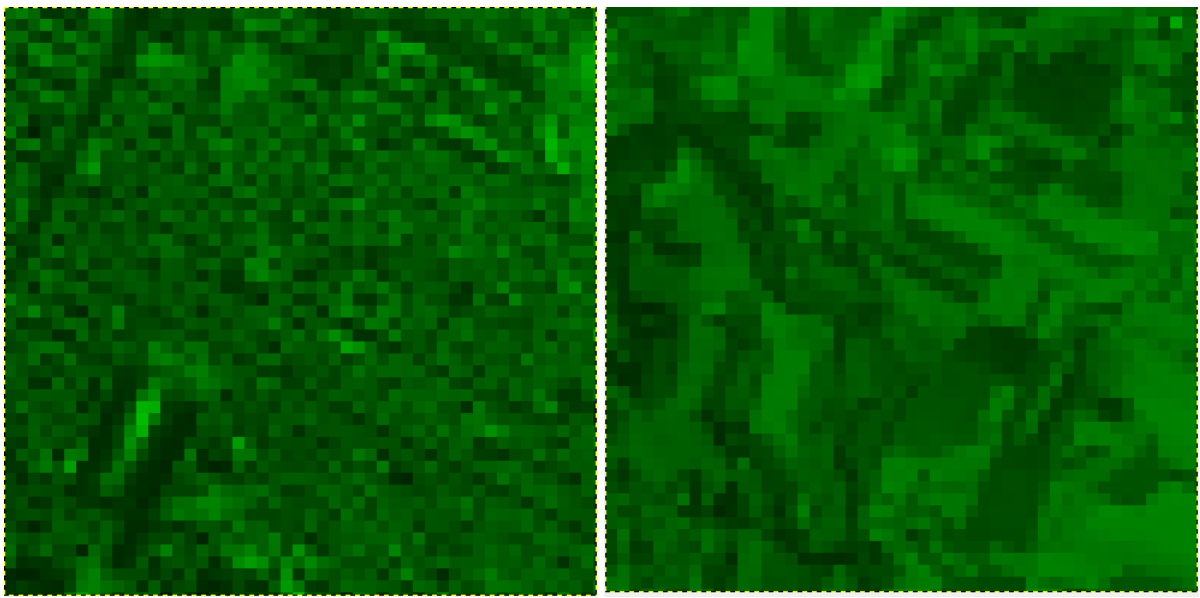

Figure 3. Pixelation of the images in the Figure 1 obtained using GIMP.

## 4. Comparing urbanization in 1945 and 2004

Now, let us apply the method for evidencing the evolution of the urbanization through the satellite images. For the analysis of the urban evolution we consider a location in England. The town is Carterton, which is today the second largest town in West Oxfordshire. The 2011 Census recorded the parish's population as 15,769 [Wikipedia]. In 1945, the town was smaller than today, as we can see in the aerial image of 1945 (see Figure 4).

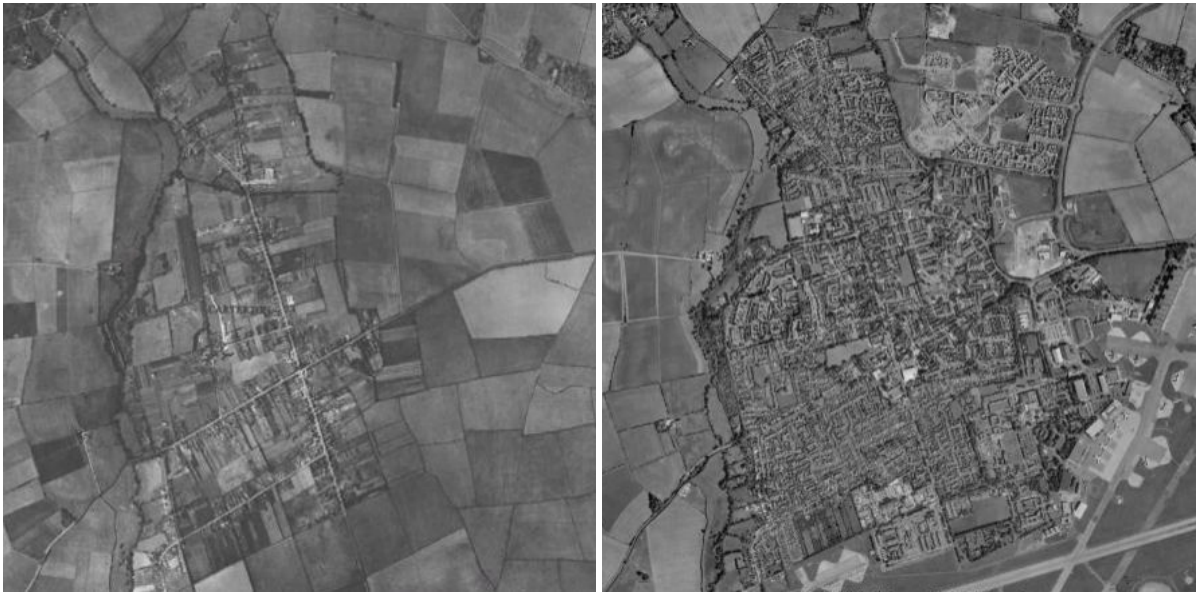

Figure 4: In the left panel we have Carterton in an aerial image of 1945, and, in the right panel, the town in a satellite image of 2004 (Courtesy Google Earth).

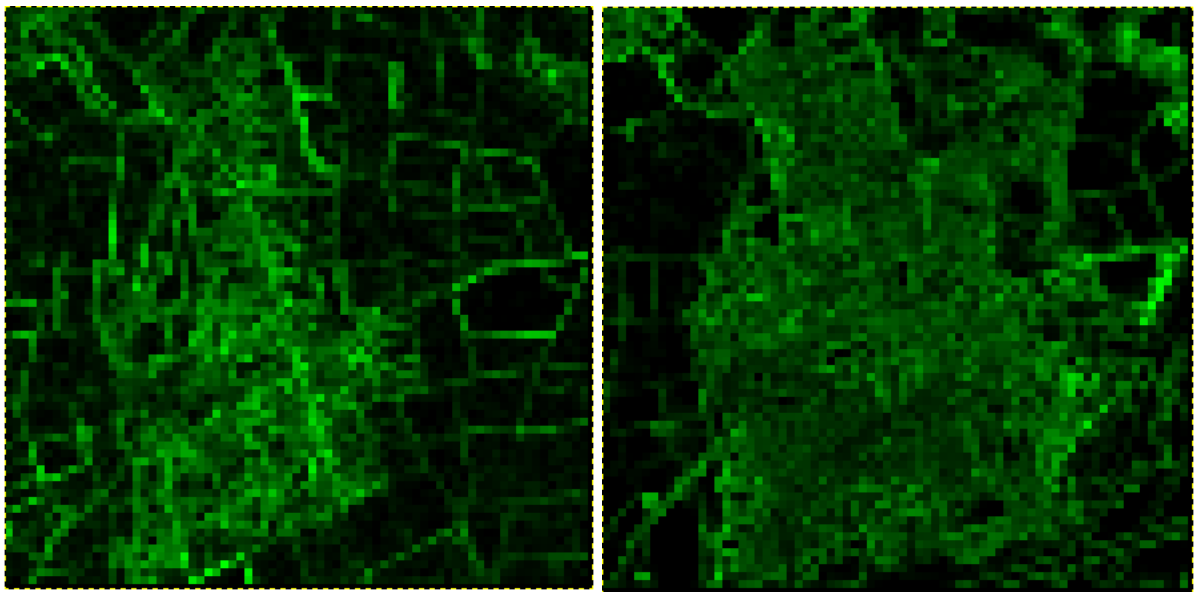

Figure 5: These two maps, obtained by the statistical approach given in the section 2, are showing that the urbanization changed. From the map on the right, we deduce that the original image has a large part of it which is containing local oscillations of the grey tones, oscillations which correspond to the presence of houses and streets. These oscillations are increasing the standard deviation of the squares. The areas in the map which are black correspond, in the original images, to areas covered by a homogenous color tone, that is, to the rural fields.

In the Figure 4 we have two grey tone images of the same town in England recorded in 1945 and in 2004. The image on the left is an aerial image and the image on the right a satellite image. We can apply the approach discussed in the section 2, for determining the maps of squares. The result is given in the Figure 5 and shows that the urbanization changed. From the map on the right of the Figure 5, we deduce that the original image has a large part of it which is containing local oscillations of the grey tones, oscillations which correspond to the presence of houses and streets. The areas in the map which are black correspond, in the original images, to the fields of the rural land.

In the Figure 6 we see another example. In the images we have Witney, town on the River Windrush in Oxfordshire, as it was in 1945 and in 2004. In the lower panels we have the maps of squares.

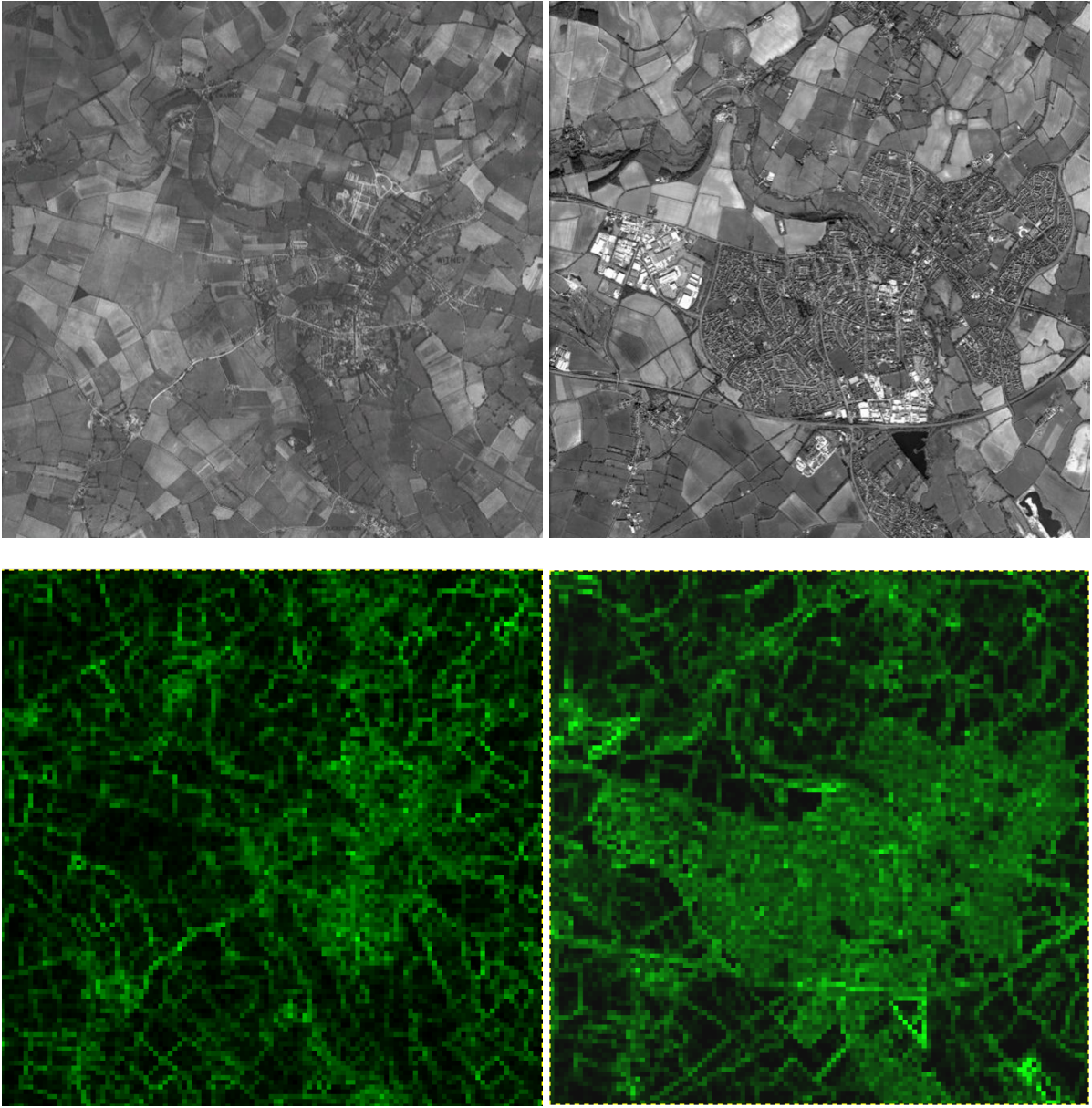

Figure 6. The same as in the Figures 4 and 5, in the case of Witney.

## 4. Texture transition

Let us consider a sequence of images recorded in 2004 of Carterton and its rural area (Figure 7). We are moving eastwards on Google Earth. We can easily observe in them a "texture transition", that is a transition from the "rural" texture to the "urban" texture. In the Figure 7 we are giving just three images to represent a sequence that can be very long, containing several images.

In 7.a we have a map characterized by black squares; in 7.c we have a large number of bright pixels. And 7.b is between them. We can tell that we have a texture transition, because 7.a has a texture different from 7.c (for the texture transitions of the liquid crystals, see please [1-3]).

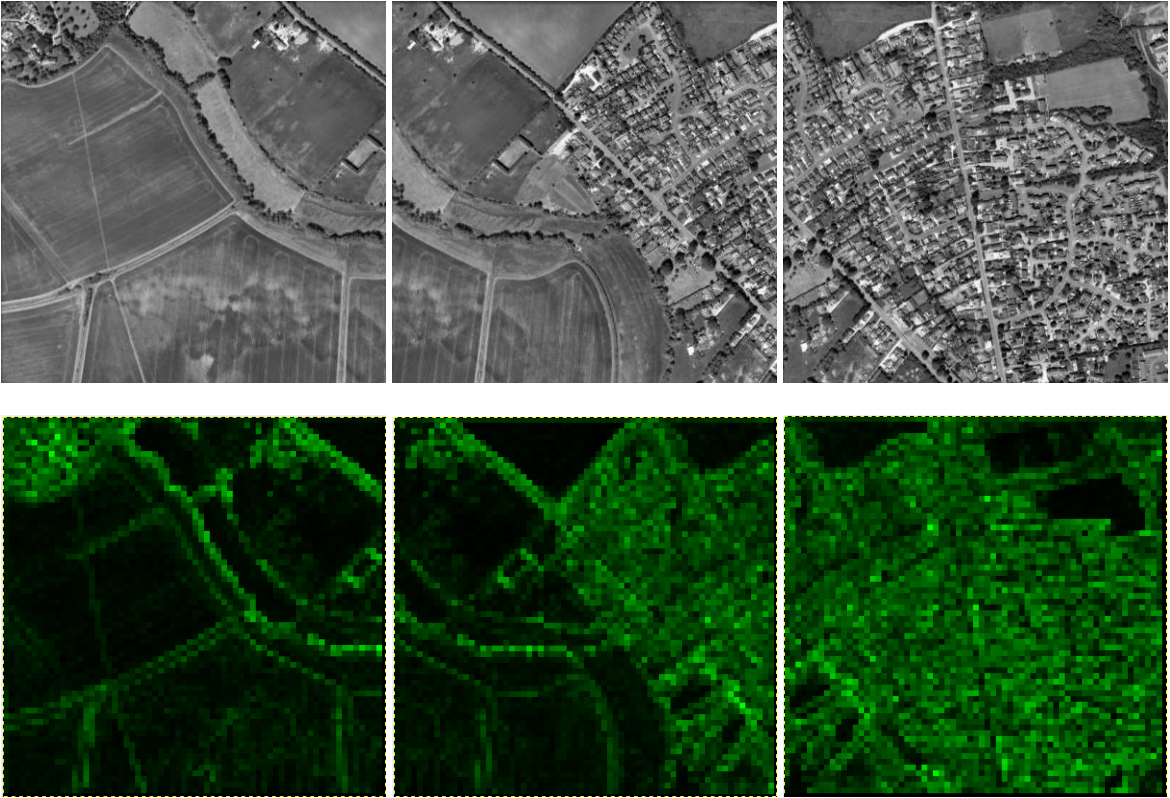

Figure 7: Sequence of images of 2004; from left to right, 7.a, 7.b, and 7.c.

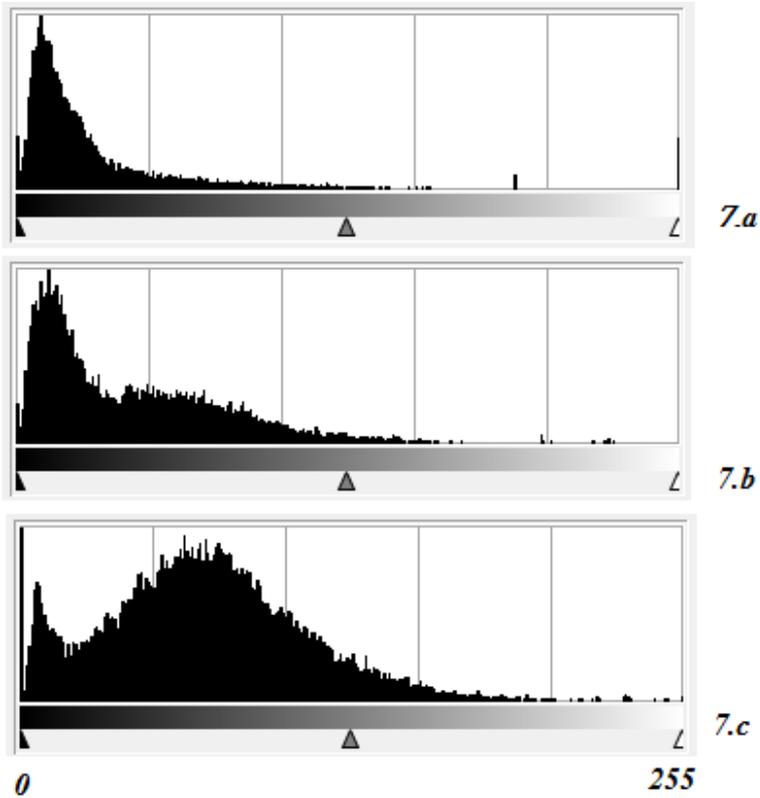

Figure 8. Histograms of the maps.

The textures are different because they have different histograms of their colour tones (the histograms are given in the Figure 8). The rural texture is characterized by a peak near low values of color tones, whereas the urban texture has two peaks, one is larger and broad corresponding to higher values of color tone. The texture transition is therefore a transition from an unimodal histogram to a bimodal histogram.

**5. Conclusion**
The aim of the paper is that of evidencing the processes of urbanization from the Google Earth time series of images. The method is a statistical method based on mean values and standard deviations, calculated from the pixels of satellite or aerial images. Using these quantities, a map is obtained, from which it is possible to distinguish urban from rural areas. The method can be applied to other processes that we can observed in satellite images.